%% file: main.tex
\pdfoutput=1

\documentclass[11pt]{article}

\usepackage[preprint]{acl}

\usepackage{times}
\usepackage{latexsym}

\usepackage[T1]{fontenc}

\usepackage[utf8]{inputenc}

\usepackage{microtype}

\usepackage{inconsolata}

\usepackage{listings}
\usepackage{xcolor}
\usepackage{float}

\definecolor{codegreen}{rgb}{0,0.6,0}
\definecolor{codegray}{rgb}{0.5,0.5,0.5}
\definecolor{codepurple}{rgb}{0.58,0,0.82}
\definecolor{backcolour}{rgb}{0.95,0.95,0.92}

\lstdefinestyle{mystyle}{
    backgroundcolor=\color{backcolour},   
    commentstyle=\color{codegreen},
    keywordstyle=\color{magenta},
    numberstyle=\tiny\color{codegray},
    stringstyle=\color{codepurple},
    basicstyle=\ttfamily\footnotesize,
    breakatwhitespace=false,         
    breaklines=true,                 
    captionpos=b,                    
    keepspaces=true,                 
    numbers=left,                    
    numbersep=5pt,                  
    showspaces=false,                
    showstringspaces=false,
    showtabs=false,                  
    tabsize=2
}

\newfloat{lstfloat}{htbp}{lop}
\floatname{lstfloat}{Algorithm}

\lstdefinestyle{myverbatim}{
    basicstyle=\ttfamily\footnotesize,
    backgroundcolor=\color{white},
    breaklines=true,
    breakatwhitespace=true
}

\usepackage{verbatimbox}
\usepackage{graphicx}
\usepackage{booktabs}
\usepackage{array}
\usepackage{algorithm}
\usepackage{algpseudocode}

\usepackage{booktabs}
\usepackage{colortbl}

\definecolor{lightgreen}{rgb}{0.9, 0.99, 0.9}
\definecolor{darkgreen}{rgb}{0.1, 0.93, 0.1}

\usepackage{graphicx}
\usepackage{multirow}
\usepackage[para]{footmisc}

\title{Poetry2Image: An Iterative Correction Framework for Images Generated from Chinese Classical Poetry}

\author{Jing Jiang$^{1}$\footnotemark[1], Yiran Ling$^1$\footnotemark[1], Binzhu Li$^1$\footnotemark[1], Pengxiang Li$^1$\footnotemark[1], Junming Piao$^1$, Yu Zhang$^{1,2}$\footnotemark[2]\\
{$^1$Harbin Institute of Technology}\\
{$^2$Research Center for Social Computing and Information Retrieval (SCIR)}\\
{\texttt{ \{2021110679,2021110742,2021112888,2021110869,2022111726\}@stu.hit.edu.cn}}\\ 
{\texttt{ \{zhangyu@ir.hit.edu.cn\}@hit.edu.cn}}
}

\begin{document}
\maketitle
\begin{abstract}
Text-to-image generation models often struggle with key element loss or semantic confusion in tasks involving Chinese classical poetry.
Addressing this issue through fine-tuning models needs considerable training costs. Additionally, manual prompts for re-diffusion adjustments need professional knowledge. 
To solve this problem, we propose Poetry2Image, an iterative correction framework for images generated from Chinese classical poetry. 
Utilizing an external poetry dataset, Poetry2Image establishes an automated feedback and correction loop, which enhances the alignment between poetry and image through image generation models and subsequent re-diffusion modifications suggested by large language models (LLM). 
Using a test set of 200 sentences of Chinese classical poetry, the proposed method--when integrated with five popular image generation models--achieves an average element completeness of 70.63\%, representing an improvement of 25.56\% over direct image generation. In tests of semantic correctness, our method attains an average semantic consistency of 80.09\%. 
The study not only promotes the dissemination of ancient poetry culture but also offers a reference for similar non-fine-tuning methods to enhance LLM generation.
\end{abstract}
\footnotetext[1]{*Equal contribution.} \footnotetext[2]{$^\dagger$Email corresponding.}
\input{text/intro}
\input{text/related_works}
\input{text/method}
\input{text/experiment}
\input{text/dis_lim_con}
\bibliography{bibliography}
\newpage
\input{text/appendix}

\end{document}

%% file: text/intro.tex
\section{Introduction}

\begin{figure}[t!]
	\includegraphics[width=\columnwidth]{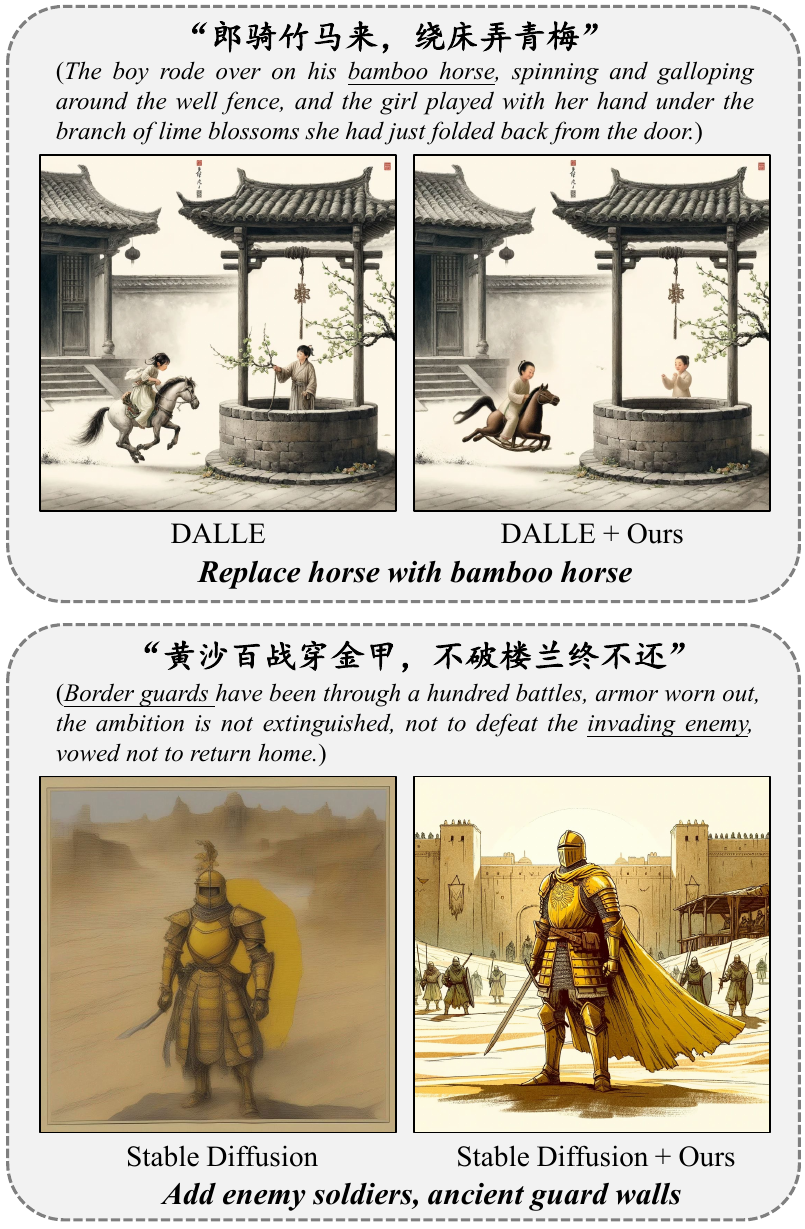}
	\caption{Direct text-based image generation often results in losing key elements in the image. Our method addresses this issue by implementing targeted image corrections, effectively capturing the semantics and artistic essence conveyed by the poem.}
	\label{fig:kaitou}
    \vspace{-11pt}
\end{figure}

Text-to-image generation combines natural language understanding with image generation models, which synthesize realistic images conditioned on natural language descriptions.
When text-to-image generation models deal with prompts requiring professional knowledge, such as Chinese classical poetry, they are prone to losing key elements or causing semantic confusion. 
It is challenging to accurately describe the precise meaning of poetry as illustrated in Fig. \ref{fig:kaitou}.

Some existing works have made efforts to alleviate these problems. 
One solution \citep{avrahami2022blended,hertz2022prompt} focuses on image editing to refine the generated images, but they suffer from complicated prompts or image understanding.
Researchers also use Lora lightweight fine-tuning \citep{hu2021lora}, retrieval-augmented generation from external knowledge database \citep{gao2023retrieval}, and specialized poetry models such as Jiuge \citep{zhipeng2019jiuge,deng2020iterative,yi2020mixpoet} to construct poetry-specific models. 
These methods, however, result in additional training costs and limited compatibility between models. 

\textit{Can external knowledge database be incorporated to edit the generated images and alleviate inconsistency between poetry and image?}

In this work, we introduce Poetry2Image, an iterative correction framework for image generation from Chinese classical poetry. 
This method identifies key elements in the initial generated image and employs text-guided image editing to alleviate inconsistency between poetry and image.
Distinct from the conventional open-loop generation approach, our method presents a closed-loop generation process capable of iteratively refining the initial image.

Initially, the retrieval system searches the input poetry in the poetry database and returns its translation and appreciation. 
Subsequently, an initial image is generated from the translation. 
The large language model (LLM) extractor is then employed to extract key elements. 
The initial image and key elements are simultaneously input to the Open Vocabulary Detector to obtain information about elements in the initial image. 
Through the element information, the LLM suggester provides modification suggestions presented as a box selection in the image. 
Image editing models apply these suggested modifications to edit the initial image.
Finally, the process above will be iterated multiple times to improve the consistency between poetry and image till no more suggestions.

Notably, Poetry2Image has no constraints on text-to-image generation models utilized for initial image generation. 
Furthermore, iterative correction operations eliminate the need for additional training costs, while the automated image generation and feedback process significantly reduces manual annotation. \textbf{The main contributions of this study can be summarized as follows:}

1. We introduce an Iterative Correction Framework for images generated from Chinese classical poetry, alleviating the loss of key elements and semantic confusion. 

2. The proposed method is not only compatible with mainstream text-to-image generation models (e.g. DALL-E) but also training-free.

3. We discuss the generalization capabilities and limitations of adopting external knowledge databases for image generation, which provides a reference for similar non-fine-tuning methods to enhance LLM generation.

\begin{figure*}[t]
	\includegraphics[width=\textwidth]{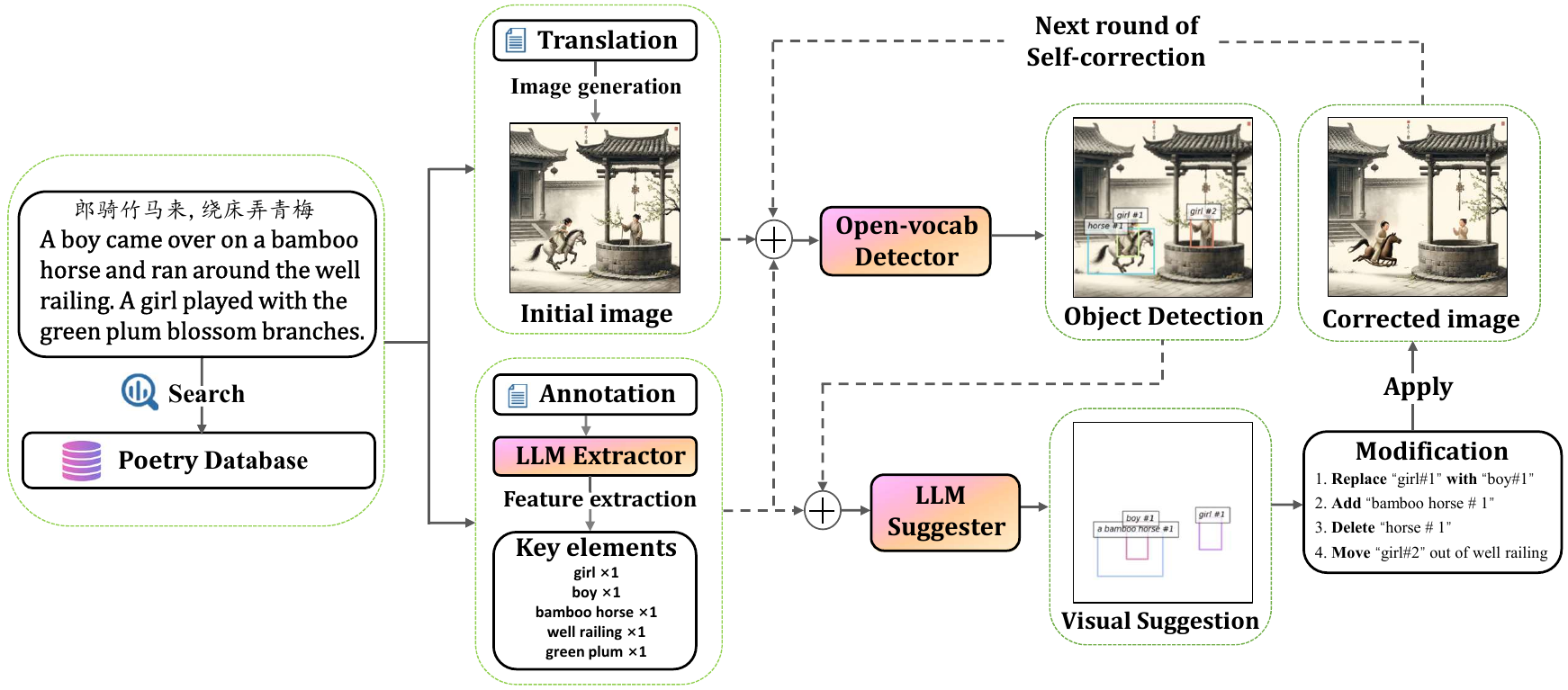}
	\caption{Automated iterative correction framework for images generated from poetry. Utilizing a pre-built poetry dataset, the process begins with the extraction of the poetry and generation of an initial image, followed by the implementation of a self-feedback image correction iteration loop. The loop functions by analyzing the semantics of the poem text and the image elements identified by Open Vocabulary Detector (OVD), utilizing LLM. It then outputs correction suggestions that guide the diffusion models for image editing, continuously providing feedback to progressively align the text semantics with the image semantics.}
	\label{fig:framework}
\end{figure*}

%% file: text/related_works.tex
\section{Related Works}

\subsection{Text-to-Image Generation}
Text-to-image generation is the task of synthesizing images conditioned on natural language prompts. 
Recent advancements in diffusion models \citep{sohl2015deep,dhariwal2021diffusion,song2020score} have significantly improved the quality of text-to-image generation, such as Dreambooth \citep{ruiz2023dreambooth} and DALL-E 3 \citep{betker2023improving}. 
Despite their impressive visual quality, these models struggle with complex prompts, which tend to generate images lacking core semantic elements and cause semantic confusion \citep{feng2022training,lian2023llm,bar2023multidiffusion}. 
Some recent studies \citep{xie2023boxdiff,yang2023reco,lian2023llm} incorporate bounding boxes as conditional controls to the diffusion process. 
Several recent papers \citep{huang2023composer,xu2024imagereward,fang2023realigndiff} leverage image understanding feedback, which builds a general-purpose reward model to refine diffusion models for text-image alignment. 
Despite their progress, there are two limitations in handling image generation with complex prompts: 
(i) open-loop generation in a single iteration cannot guarantee the alignment between generated images and prompts; 
(ii) these methods result in additional training costs. 
To address these issues, we introduce a training-free cyclic self-correction framework to enhance the alignment of images with complex prompts.

\subsection{Text-Guided Image Editing}
Text-guided image editing synthesizes images from a given image and text descriptions. 
Classic image editing aims at fine-grained manipulation by inpainting masked regions while keeping the remaining areas. 
Studies \citep{avrahami2022blended,meng2021sdedit} show that using user-generated masks for spatial editing in image generation is a straightforward yet effective method. 
Another method \citep{balaji2022ediff,hertz2022prompt} focuses on predicted masks for spatial editing, demonstrating that manipulating image-text cross-attention masks is also effective. 
Evolved from spatial editing, text-guided image editing \citep{brooks2023instructpix2pix,kawar2023imagic} accepts direct commands, allowing editing without regional masks. 
Despite some progress, these works mainly focus on diffusion models but often suffer from complicated prompts or image understanding. 
Recent advancements have demonstrated the capabilities of incorporating external language \citep{brooks2023instructpix2pix} or vision \citep{kirillov2023segment} pre-trained models for editing. 
These methods, however, struggle with fine-grained manipulation according to user-provided texts when editing images in a single iteration.

%% file: text/method.tex
\section{Method}

In this section, we introduce the iterative correction framework for poetry generation, as shown in Fig.~\ref{fig:framework}. 
Compared to common texts, poetry is semantically implicit. 
Firstly, the implicit semantic elements should be extracted.
Secondly, an image semantic error correction mechanism should be established to alleviate potential semantic inconsistencies in the images. 

\subsection{Extract Implicit Semantics Based on LLM}
\label{sec:3.1}

\textbf{Dataset construction} should consider the meanings of poems, completeness of translation annotations, and cultural popularity. 
We consider rhetorical techniques involved in literature with semantic implicit features such as metaphor, personification, hyperbole, and allusion.
Then, we allocate 200 well-known sentences with their modern Chinese translations, keyword annotations, and phrase explanations from the largest public platform of Chinese classical poetry, GuShiWen.com\footnote{https://www.gushiwen.cn/}.

\begin{algorithm}[h]
	\caption{\textbf{Key Elements Extraction}}
	\label{alg:algorithm1}
	\begin{algorithmic}[1] 
		\Require{Poetry $p$; Poetry Database $S$}
		
		\State $d_{min} \gets 0$
            \State $p_{find} = \emptyset$ // Query list initialization
		\For{$i = 1$ to $N$}
		\State $d = F_{simularity}(p, S[i])$
		\If{$d \leq d_{min}$}
		\State $d_{min} \gets d$
		\State $p_{find}.append(S[i])$
		\EndIf
		\EndFor
		\State $t_{find} = F_{Translation}(p_{find})$
		\State $n_{find} = F_{Annotation}(p_{find})$
		\State $E_{key} = {LLM_{extract}}(p_{find}, t_{find}, n_{find})$
		\State \textbf{Initial image}: $P_{origin} = {Diff}(t_{find})$
		
		\Ensure{Key elements $E_{key}$; Initial image $P_{origin}$}
	\end{algorithmic}
\end{algorithm}

\begin{figure}
	\includegraphics[width=\columnwidth]{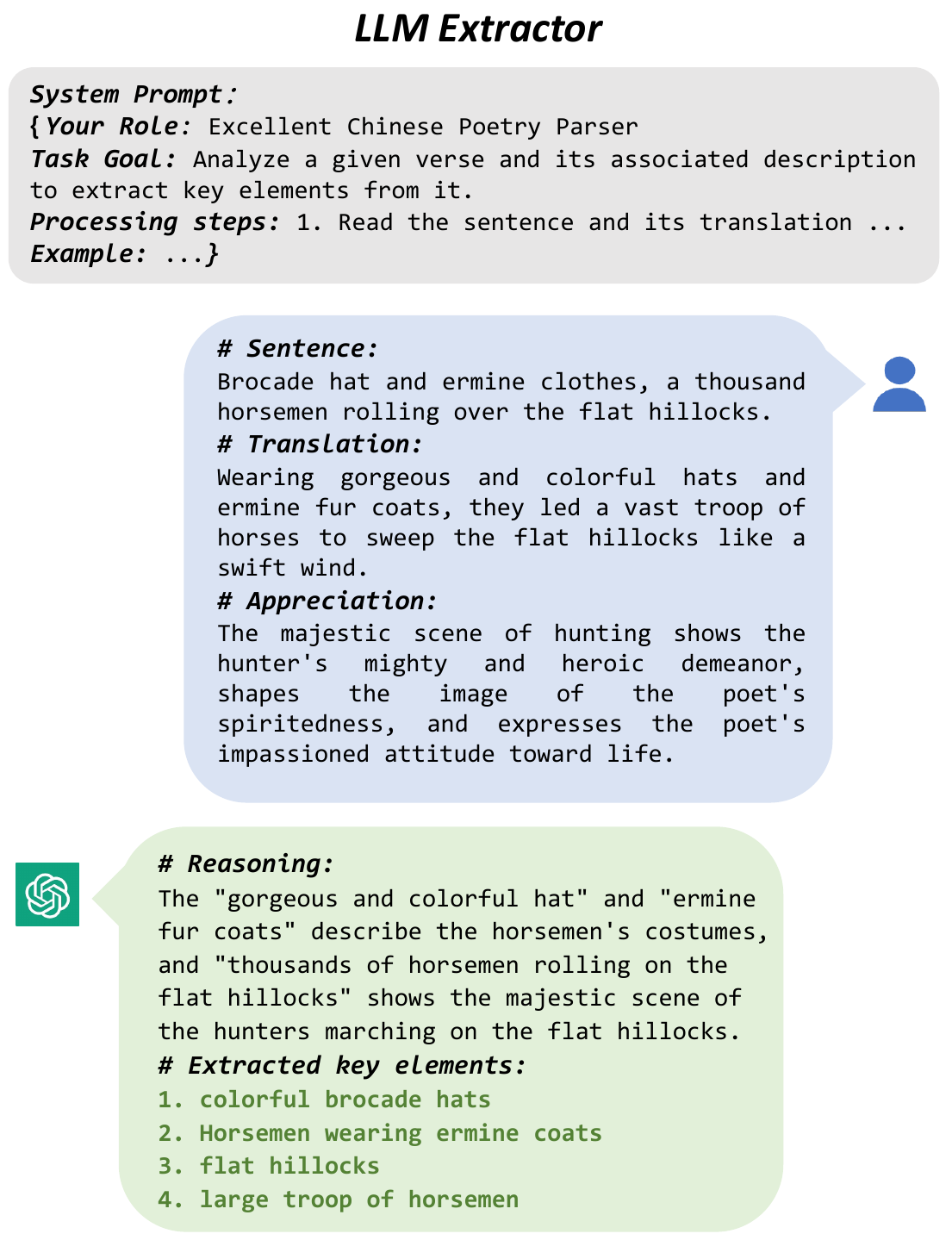}
	\caption{An illustration of the LLM Extractor, a key element extraction module. Upon retrieving the poem's translation and critical appreciation from the poetry database, these texts along with the system prompt are fed into the LLM. Subsequently, the LLM outputs the key elements contained in the poetry.}
	\label{fig:extractor}
\end{figure}

\noindent \textbf{Data extension} involves processing the dataset to match the input features required for detection and ensuring generality for prompts from different image generation models. Key elements of the poetry are extracted along with their translations, appreciations, and annotations, to facilitate monitoring of the elements completeness in the generated images. To automate the extraction process and achieve high extraction accuracy, we use GPT-4 for key element extraction, and design prompts for the LLM, as illustrated in Fig. \ref{fig:extractor}. 

We use the extracted key elements as supervisory text for subsequent image editing. The procedure is shown in Algorithm~\ref{alg:algorithm1}.

$P$ denotes collections of images, distinguished by subscripts. $d$ measures semantic distance, ranging from 0 to 1. $F_{similarity}$ calculates semantic cosine similarity. $F_{translation}$ searches for poetry translations, denoted by $t_{find}$. $F_{Annotation}$ searches for poetry annotations, denoted by $n_{find}$.

Additionally, we conduct manual element extraction on the poem and compare these results with those extracted by LLMs to validate the effectiveness of our methodology.

\subsection{Automated Iterative Correction Framework}
\label{sec:3.2}

\textbf{Initial image generation} focuses on using translations of poetry as inputs for generating images instead of original poems. This approach ensures the images accurately reflect the poems' meanings, avoiding ambiguities caused by historical linguistic changes and complex rhetorical devices such as metaphors and personifications. 

\begin{figure}[h!]
	\includegraphics[width=\columnwidth]{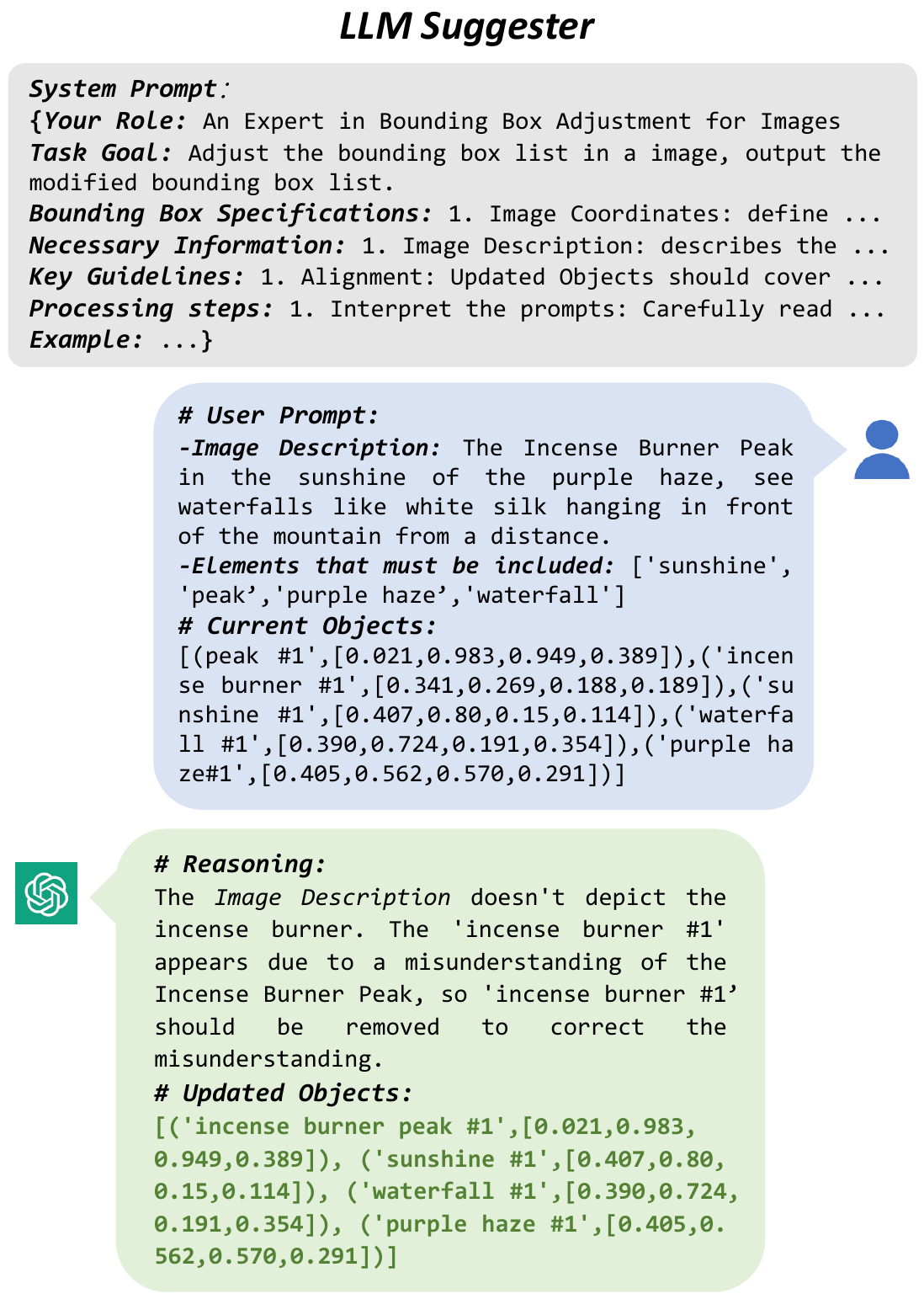}
	\caption{An example of the LLM Suggester, a module dedicated to modifying image bounding boxes. After conducting OVD-based element recognition to determine the existing bounding box, the translation, this bounding box, and the system prompt are inputted into the LLM. The LLM then adjusts the bounding box based on the semantic information in the translation, outputting the modified bounding box.}
	\label{fig:suggester}
\end{figure}

\noindent \textbf{Detecting and correction} involves identifying where the key elements of the image are located. 
We use the Open Vocabulary Detector (OVD), an open-source recognition method based on an open corpus, to build the front part of our image correction component. 
The input to this part includes the initial generated images, and the recognition labels derived. 
After the OVD performs these extractions, feedback suggestions on the bounding boxes is generated, which will be transmitted to LLM for analysis in the form of labels and region annotations, as illustrated in Fig. \ref{fig:suggester}. 
The LLM suggester provides modification suggestions and proposes a new box for the image elements. 
The extracted elements need to be compared with the labels of the elements in the bounding box to detect whether complete and correct elements are in the initial generation of the diagram. 
Algorithm~\ref{alg:algorithm2} shows the procedure in detail.

\begin{algorithm}
	\caption{\textbf{Image Feedback Correction}}
	\label{alg:algorithm2}
	\begin{algorithmic}[1] 
		\Require{Key elements $E_{{key}}$; Initial image $P_{origin}$}
		
		\State \textbf{key elements detection: }$E' \gets \emptyset$
		\State $P_{feedback} \gets P_{origin}$
		
		\While{$E' \neq E_{{key}}$}
		\State $L' = {OVD}(P_{feedback})$
		\State $L'' = {LLM_{suggester}}(L', E_{{key}})$
		\State $L''' = {LLM_{transform}}(L'')$
		\State $P_{feedback} = Diff(L''')$
		\State $E' \gets {OVD}(P_{feedback})$
		\EndWhile
		
		\State \textbf{return} $P_{out} \gets P_{feedback}$
		
		\Ensure{Final generated image $P_{out}$}
	\end{algorithmic}
\end{algorithm}

$L$ represents a list of intermediate calculation results. $E$ represents a list of key elements in the semantics of poetry.

For all the bounding boxes in the image, the element labels are compared with the results from the LLM Extractor to determine whether a bounding box need to be retained or modified. This is discussed in the following scenarios:

1. Retain: Keep the bounding box unchanged if it is included in the LLM Extractor's result, which is key elements.

2. Remove: Delete the bounding box based on the LLM analysis of the poetic imagery.

3. Add: If an key element from the LLM Extractor's result is missing in the current generated image, the LLM selects a new rectangular area and adds the missing key element label based on the poetry translation.

4. Move: If there is a positional conflict between bounding boxes in the current generated image, the LLM selects a new position, deletes the bounding box in the origin position, and regenerates it in the new position.

5. Replace: If a bounding box in the current image conflicts with the LLM Extractor's result, the LLM deletes the element and adds a proper element in the original position.

After prompt generation, we obtain suggested modifications, which can simplify the issue into a standard text-based editing task. We select the appropriate open-source diffusion models, input the suggested modifications, and use the bounding boxes and labels from the LLM suggester to guide SAM 
\citep{kirillov2023segment} for semantic segmentation, completing a round of image modifications.

\noindent \textbf{Recycle and Finish} involves determining if the image correctly contains all key elements. We first detect all elements in the initial correction images and generate border images. These border images and element words are then input into the LLM Suggester. If the LLM Suggester provides new modification suggestions, they are applied to generate subsequent correction images, and the process repeats. If no new suggestions are provided or the loop reaches the preset limit, the image and text are deemed consistent, and the loop exits, yielding the final result.

\noindent \textbf{Evaluation} involves the elemental completeness and the semantic consistency in the poetry generated image, and we establish an image-text consistency evaluation model, as shown in Eq.~\ref{equation}.

\begin{equation}
\label{equation}
\arg\max_{s,e} \Theta = \frac{\alpha \left(\frac{S}{s_{\epsilon}}\right) + \beta \left(\frac{e}{e_{\epsilon}}\right)}{\alpha + \beta}
\end{equation}

$\Theta$ is the quantitative measure for assessing generated images of ancient poems, considering semantic features $s$ and key elements $e$, with upper thresholds $s_{\epsilon}$ and $e_{\epsilon}$ respectively. Linear parameters $\alpha$ and $\beta$ dictate the focus: $\beta=0$ evaluates semantic compliance, while $\alpha=0$ evaluates key elemental completeness.

\begin{table}[h]
	\centering
	\begin{tabular}{lcc}
		\toprule
		\textbf{Model} & \textbf{Average Similarity} & \textbf{Rank} \\ \midrule
		GPT-4-Turbo & 0.8740 & 4 \\ 
		GLM-4 & 0.8763 & 3 \\ 
		\textbf{Claude-3} & \textbf{0.8868} & \textbf{1} \\ 
		GPT-3.5-Turbo & 0.8660 & 5 \\ 
		ERNIE-4.0 & 0.8834 & 2 \\ \bottomrule
	\end{tabular}
	\caption{Evaluation of Element Extraction Effectiveness of Various Large Language Models. According to this evaluation, the Claude-3 model exhibits the highest effectiveness in key element extraction.}
	\label{tab:performance}

\end{table}

%% file: text/experiment.tex
\section{Experiment}
\begin{table*}[h!]
	\centering
	\begin{tabular}{lll}
		\toprule
		\textbf{Method} & \textbf{Elemental Completeness} & \textbf{Semantic Consistency}\\
		\midrule
		DALL-E & 56.33\% & 81.94\% \\
        \rowcolor{lightgreen}
		DALL-E+\textbf{ours}  & 90.20\% \textbf{(+33.87\%)}& 84.18\% \textbf{(+2.24\%)} \\
		\midrule
		CogView  & 50.69\% & 77.77\% \\
		CogView+\textbf{ours}  & 67.28\% (+17.59\%) & 78.82\% (+1.05\%)\\
		\midrule
		Wenxin Yige  & 33.17\% & 80.95\% \\
		\rowcolor{lightgreen}
		Wenxin Yige+\textbf{ours}  & 64.76\%  \textbf{(+31.58\%)} & 81.77\%  (+0.82\%)\\
		\midrule
		Stable Diffusion  & 37.71\% & 72.25\% \\
		Stable Diffusion+\textbf{ours}  & 63.12\%  (+25.41\%) & 73.87\%  \textbf{(+1.62\%)}\\
		\midrule
		Midjourney & 48.45\% & 80.06\% \\
		Midjourney+\textbf{ours}  & 67.78\% (+19.33\%) & 81.79\% \textbf{(+1.73\%)} \\
		\bottomrule
	\end{tabular}
	\caption{Comparison with Image Generation Models. Our method shows a significant improvement in elemental completeness through image generation models. For elemental completeness, the accuracy improvement ranges from 17.59\% to 33.87\%, and for semantic consistency, it also achieves a certain degree of improvement..}
	\label{tab:comparison}
\end{table*}
\subsection{Key Elements Extraction}
In our image generation process, the initial stage uses LLM Extractor to semantically extract key elements from the database corpus. The accuracy of LLM Extractor is crucial to the subsequent process and needs to be evaluated in detail.

\noindent\textbf{Settings.} We select a dataset of 200 Chinese poems with implicit semantics and manually annotated them to establish a benchmark for element extraction. The poems are then processed using five LLMs: GPT-4-Turbo \citep{achiam2023gpt}, GPT-3.5-Turbo \citep{brown2020language}, Claude-3 \citep{anthropic2024claude}, GLM-4 \citep{zeng2022glm}, and ERNIE-4.0 \citep{sun2019ernie}. To assess the effectiveness of key element extraction by these LLM Extractors, the BERT-based-Chinese model \citep{devlin2019bert} was employed. The similarity between the manually annotated key elements and elements extracted by the LLMs served as a quantitative performance indicator.

\noindent\textbf{Results.} Based on the cosine similarity evaluation, effectiveness scores for element extraction were shown in Tab. \ref{tab:performance}. In terms of semantic understanding, Claude-3 exhibited the highest performance with a score of 0.8868, closely followed by ERNIE-4.0. GPT-4-Turbo and GLM-4 demonstrate comparable performance, whereas GPT-3.5-Turbo shows marginally reduced accuracy. Given its superior performance, Claude-3 is also employed as the LLM Extractor in experiments that do not involve LLM tuning. Overall, the five LLMs tested within  our method achieve an element extraction accuracy exceeding 0.85, demonstrating notable consistency. Therefore, we contend that our method substantiates the use of LLMs for extracting key elements in Chinese poems, providing a robust foundation for subsequent processes.

\begin{figure*}
	\includegraphics[width=\textwidth]{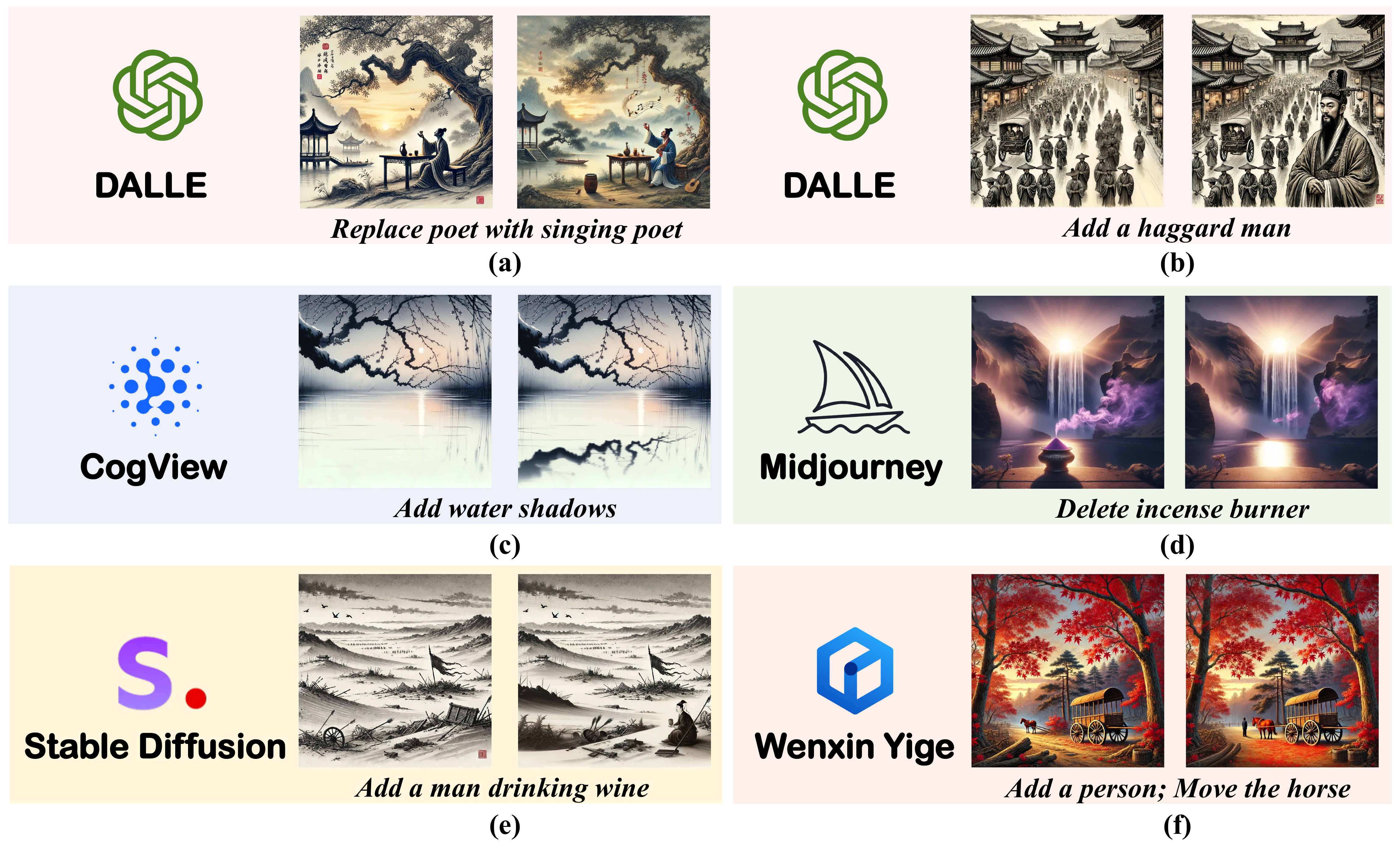}
	\caption{Image generation effect of the whole process evaluation. Peotry2Image enhances image generation quality for specialized texts like classical poetry and addresses core issues such as morpheme loss and semantic confusion. The poems corresponding to the images can be found in Appendix \ref{sec:B}.}
	\label{fig:datu}
 \vspace{-5mm}
\end{figure*}

\subsection{Verification of Self-Correcting Cycles Across Different Generation Models}

\textbf{Settings.} We evaluate our image error correction method using elemental completeness and semantic consistency. Poetry2image is applied to five text-to-image generation models to validate the effect: DALL-E-3 \citep{peebles2023scalable}, CogView3 \cite{zheng2024cogview3}, Midjourney, Wenxin Yige, and Stable Diffusion \citep{esser2024scaling}. Each model is assessed using a dataset of 200 Chinese poems with complex semantics, as shown in Fig. \ref{fig:datu}. Utilizing Open-vocabulary Detector OWL-ViT v2 \citep{minderer2024scaling}, we quantify key elements in both the initial and corrected images to determine elemental completeness. Employing BERT-Chinese \citep{devlin2019bert}, we measure semantic consistency by comparing the image content with the corresponding translation.

\noindent \textbf{Result.}  The full-process evaluation results are shown in Tab. \ref{tab:comparison}. In the key elemental completeness test, Peotry2Image achieved an accuracy improvement ranging from 17.59\% to 33.87\%. Moreover, the elemental completeness of DALL-E has reached 90.20\%, demonstrating good performance. In the semantic correctness test, the average semantic consistency reached 81.64\%. Peotry2Image maintained stability in semantic consistency, indicating that it meets the standards for consistency between image content and poem sentiment. This stability also suggests that the improvement in image quality primarily results from enhanced element integrity.

\subsection{Comparison of Iteration Rounds}

\textbf{Settings.} To ensure the improvement in elemental completeness and ascertain the maximum efficacy of our method, we conduct an iterative comparison experiment. Observation points are established within the automated process, sequentially recording improvements in elemental completeness as the number of image iteration rounds increased. 

\noindent \textbf{Result.} Based on the experimental outcomes as shown in Tab. \ref{tab:iteration round}, the following conclusions can be drawn:

1. With increasing image iteration rounds, elemental completeness improves, achieving a notable increase of \textbf{27.30\%} by the first round.

2. The elemental completeness of the images ultimately stabilizes at approximately \textbf{90\%}around 3 rounds of iterations, demonstrating our method’s effectiveness in accurately correcting and redrawing most ideal elements.
\vspace{-5mm}
\begin{table}[h]
	\centering
	\begin{tabular}{ccc}
		\toprule
		\textbf{Round} & \textbf{Elem. Completeness} & \textbf{Improv.} \\ \midrule
		0 & 56.33\% & - \\ 
		1 & 83.63\% & \textbf{+27.30\%} \\ 
		2 & 87.50\% & +3.87\% \\ 
        3 & 90.20\% & +2.70\% \\ 
		\bottomrule
	\end{tabular}
	\caption{The relationship between the number of iterations and elemental completeness shows that as the number of image iterations increases, the completeness of elements in the images correspondingly rises, achieving a significant gain of 27.30\% by the first round.}
	\label{tab:iteration round}
\end{table}

\subsection{Ablation Experiment}

\textbf{Settings.} In order to verify the effect of the initial generation on correction, we perform an ablation experiment, removing additional information such as translations and annotations, and directly utilize the original text of the poems for generation.

\noindent \textbf{Result.} The results of the experiment show in Tab. \ref{tab:Ablation experiment}. The completeness of the initial generation remains largely unchanged after eliminating additional information, while after detection and correction the elemental completeness decreases by approximately 11\%. This is because most elements are derived directly from the poem's text, so the initial generation's completeness is unaffected. However, images generated directly based on poems lack much of the additional semantics from the additional information, due to factors such as lack of stylistic richness, which reduces the completeness of the picture elements, and subsequent modifications can be much less effective.
\begin{table}[h!]
	\centering
	\begin{tabular}{ccc}
		\toprule
		\multirow{2}{*}{\textbf{Setup}} & \multicolumn{2}{c}{\textbf{Elemental Completeness}} \\ \cmidrule{2-3}
		& \textbf{Initial Image} & \textbf{First Round} \\ \midrule
		Poetry & 54.61\% & 72.50\% \\ 
		Translation & 56.33\% & 83.63\% \\ 
		\bottomrule
	\end{tabular}
	\caption{Ablation experiment result. After the elimination of the additional information, the completeness of the initial generation remains largely unchanged, while the completeness of the elements after the detection and correction decreases by approximately 11\%. }
	\label{tab:Ablation experiment}
\end{table}

%% file: text/dis_lim_con.tex
\section{Discussion}

\subsection{The Influence of the Number of Key Elements in Poetry}

To evaluate the performance of Poetry2Image in processing Chinese classical poetry with different numbers of key elements, we design a series of experiments and use elemental completeness as the evaluation indicator. 

The experimental results, as illustrated in Tab.~\ref{tab:Number of elements}, indicate that with fewer key elements, such as 3, the initial generation covers most elements, resulting in minimal improvement in overall elemental completeness. 
As the number of key elements increases, the initial generation's missing rate escalates. However, Poetry2Image compensates for this by completing elements more rapidly than they are missed, resulting in a 15\% to 20\% improvement in elemental completeness. Specifically, for information-intensive poems containing up to six elements, the elemental completeness improvement rate reaches 23.73\%. This demonstrates Poetry2Image's efficacy in improving elemental completeness.
However, when the number of elements exceeds seven, the image fails to achieve the desired elemental completeness, posing a challenge in balancing the aesthetics of the image with the improvement of elemental completeness.
\begin{table}[h]
	\centering
	\begin{tabular}{cc}
		\toprule
		\textbf{Num of Elements} & \textbf{Improvement}\\ \midrule
		3 & +13.63\%  \\ 
		4 & +15.35\%  \\ 
		5 & +18.61\%  \\ 
		\textbf{6} & \textbf{+23.73\%} \\ 
		7 & +3.57\%  \\ 
		\bottomrule
	\end{tabular}
	\caption{The impact of the number of key elements contained in poetry on the elemental completeness. Poetry2Image performs well when dealing with poetry with multiple key elements, ranging from 3 to 6.}
	\label{tab:Number of elements}
 \vspace{-5mm}
\end{table}
\subsection{The Influence of Poetry Language Types}
To further assess the generalizability and applicability of Poetry2Image, we extended its application to multilingual poetry. We test Poetry2Image on 100 classical Japanese and English poems representing diverse linguistic and cultural backgrounds. 

The results, as detailed in Appendix \ref{sec:A}, demonstrate that our method is effective not only with Chinese classical poetry but also with Japanese and English poetry.This confirms the wide applicability of Poem2Image and provides insights into generating images of multilingual poetry.

\section{Limitations}

The limitations of Poetry2Image stem from the intrinsic characteristics of poetry, as illustrated in Fig.~\ref{fig:limitation}. The poems corresponding to the images can be found in Appendix \ref{sec:B}. When the genre of the poem is lyrical or didactic, the key elements in the sentence are scarce or abstract, so both the initial image and the corrected image fail to capture the key elements used for generation, leading to unsatisfactory correction results. In addition, when dealing with proper nouns such as historical personal names (e.g.`Zhou Yu') in the poems, the elements cannot be recognized and understood by the OVD and diffusion models, resulting in suboptimal correction results.

\begin{figure}[h]
	\includegraphics[width=\columnwidth]{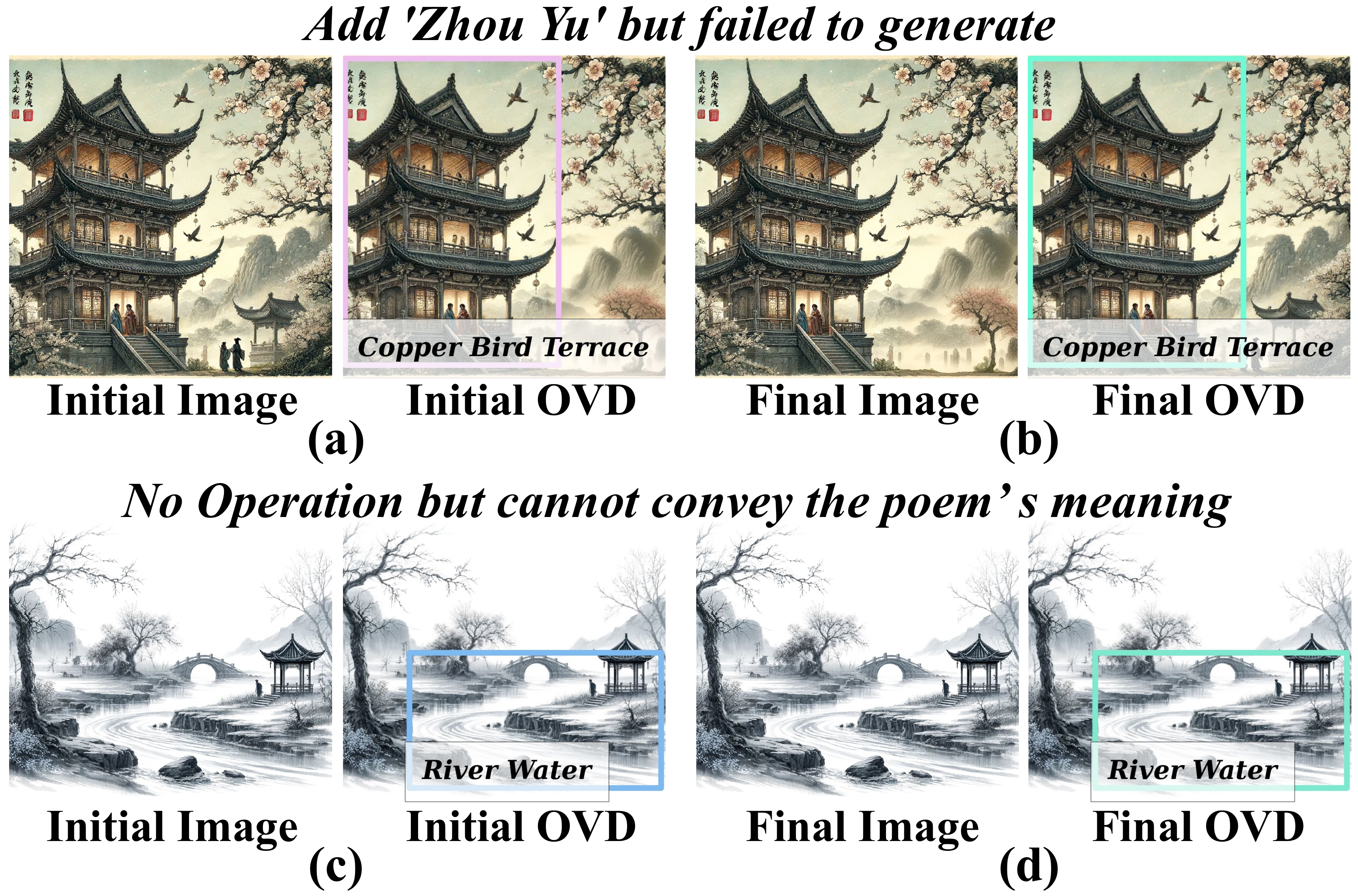}
	\caption{The diffusion model is unable to understand the key element `Zhou Yu',who is a historical figure, so cannot generate it. In the second poem, all elements can be identified, but it fails to convey the sense of nostalgia for the dead hero.}
	\label{fig:limitation}
 \vspace{-5mm}
\end{figure}

\section{Conclusion}
We propose Poetry2Image, an iterative correction framework that integrates image generation, error correction and feedback. This framework enhances image generation quality for specialized texts like Chinese classical poetry and addresses core issues such as element loss and semantic confusion. Our method is adept at element-rich or multi-lingual poems and is compatible with other image generation models.Additionally, our approach provides a reference for similar non-fine-tuning methods to enhance LLM generation.

%% file: text/appendix.tex
\appendix
\section*{Appendix}

\section{Results of Poetry Image Correction in Multiple Languages}
\label{sec:A}

Poetry examples in different languages and test results of Poetry2Image are shown below.

1. Japanese Haiku: The moon in the water; Broken and broken again, Still it is there.

2. American English Poetry: On the beach at night alone, As the old mother sways her to and fro singing her husky song, As I watch the bright stars shining, I think a thought of the clef of the universes and of the future.

3. British English Poetry: O wild West Wind, thou breath of Autumn's being Thou, from whose unseen presence the leaves dead Are driven, like ghosts from an enchanter fleeing, Yellow, and black, and pale, and hectic red.

\begin{figure}[h]
	\includegraphics[width=\columnwidth]{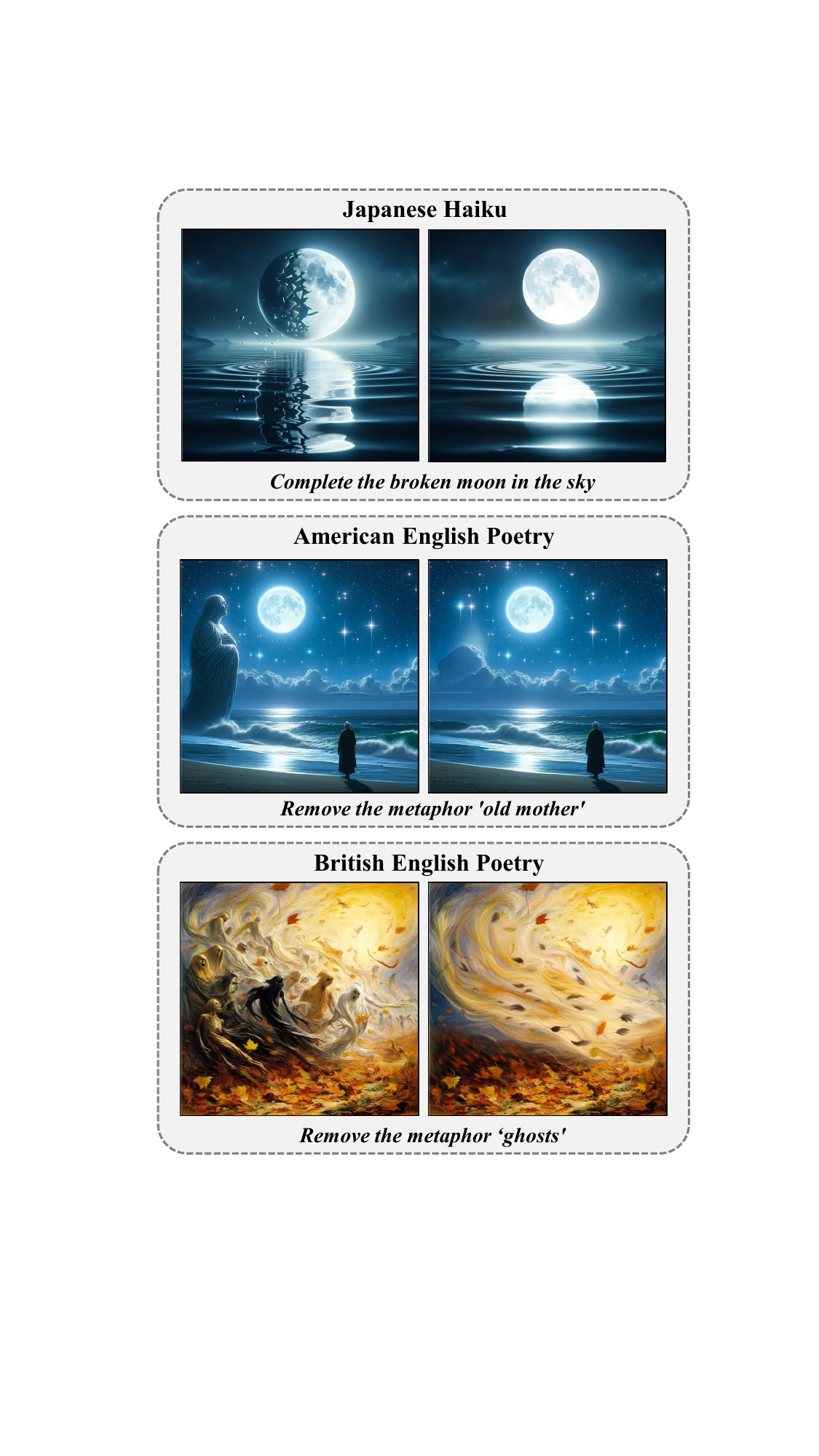}
	\caption{Poetry image generation in different languages and styles. The left is generated directly from literal meaning, and the right shows corrections of our method.}
	\label{fig:4}
\end{figure}

Initially, for Japanese poetry, we chose the renowned haiku of Matsuo Basho for analysis. Our method accurately identified the metaphor of a `broken moon in the water' and appropriately adjusted the image from a moon in the sky to reflect this. Subsequently, for English poetry, we tested poems by Whitman and Shelley. The results indicate that our method effectively interprets and corrects metaphors such as `old mother' and `ghosts'.

\section{Poetry Text for Generating Images}
\label{sec:B}
\textbf{Poetry Text for Generating Fig. \ref{fig:datu}.}

Poetry a: Singing loudly in front of the wine, life is short and the days pass by quickly.

Poetry b: The capital is filled with nobles in fine cars and beautiful clothes, but you are extremely talented but your face is haggard.

Poetry c: The sparse shadows of plum blossoms are reflected obliquely in the clear water, and the faint fragrance of plum blossoms is drifting in the hazy moonlight.

Poetry d: The Xianglu Peak is covered with purple haze under the sunlight, and from a distance you can see a waterfall hanging in front of the mountain like white silk.

Poetry e: I am facing a cup of sad wine, thousands of miles away from home. I have a lot of thoughts, thinking about the unrest on the border, the unfinished work, and I don't know when I can return to my hometown.

Poetry f: I stopped the carriage just because I loved the maple forest in the evening. The frost-stained maple leaves are more beautiful than the bright flowers in February.

\noindent\textbf{Poetry Text for Generating Fig. \ref{fig:limitation}.}

Poetry a: Without the help of the east wind, Jiangnan would have been a ruin; the beautiful Erqiao would have been locked up in the Tongque Tower forever.

Poetry b: The people back then are no longer around, but the Yishui River is still as cold today.

\section{System Prompt Setup of Extractor and Suggester in Our Method}
\label{sec:C}

We use a recognition method based on open vocabulary detector to detect key elements of poems and an automatic iterative correction framework to generate images through secondary diffusion. The system prompt setup of our extractor and suggester are shown below.

\begin{table*}[h!]
\setlength\tabcolsep{0pt}
\centering
\begin{tabular*}{\linewidth}{@{\extracolsep{\fill}} l }\toprule
\begin{lstlisting}[style=myverbatim]
# Your role: Excellent Chinese poetry parser

## Task objective: Analyze the given poem and its related descriptions, and extract the key image elements from it.

## Processing steps
1. Read the poem provided by the user and its translation and appreciation.
2. Identify all the key image elements mentioned in the poem or translation and record them. The key image elements will be used to draw an ink painting of this poem later.
3. The key image elements are listed in the form of "noun" or "adjective + noun".
4. Explain your reasoning and organize your results in the format of the example.
5. The elements must be complete, including all the key elements mentioned in the poem.
6. Abstract descriptions such as atmosphere and emotion must not appear in the key elements, such as "desolate atmosphere" and "sad mood".
7. Please ensure that the key image elements have no brackets, quotation marks or other special characters, and are specific nouns or adjective + noun combinations.

## Example
- Example 1
Original sentence: Yellow sand and golden armor worn through a hundred battles, never return until Loulan is conquered.
Translation: The soldiers guarding the border have experienced a hundred battles, their armor worn through, their ambitions undying, they will not return home until they defeat the invading enemy.
Appreciation: The first sentence shows the long time of guarding the border, the frequent battles, the hardship of the battles, the strength of the enemy, and the desolation of the border. The second sentence expresses the soldiers' lofty ambitions to serve their country to the death and their sincere patriotic enthusiasm.
Reasoning: The yellow sand and golden armor mentioned in the description reflect the hardships of border defense and the tenacity of the soldiers.
Image elements:
1. Yellow desert
2. Soldiers wearing golden armor
3. Desolate border battlefield

- Example 2
Original sentence: Brocade hats and mink furs, thousands of cavalry roll across the flat hills.
Translation: Wearing gorgeous and bright hats, wearing mink furs, leading a mighty large army, like a gust of wind, sweeping across the flat hills.
Appreciation: The magnificent scene of hunting shows the hunter's majestic and heroic spirit, shapes the poet's high-spirited image, and shows the poet's passionate attitude towards life.
Reasoning: "Brocade hat" and "sable fur" describe the cavalry's clothing, and "thousands of cavalry rolling on the flat hill" shows the magnificent scene of a large army marching and hunting on a flat hill.
Picture elements:
1. Gorgeous hat
2. Cavalry wearing sable fur
3. Broad flat hill
4. Huge cavalry team

Your current task: Follow the above steps carefully and accurately identify the screen elements based on the given poem. Be sure to follow the above output format.
\end{lstlisting} \\\bottomrule
\end{tabular*}
\caption{System Prompt Setup of Extractor in Our Method.}
\label{tab:extractor_prompt}
\end{table*}

\begin{table*}[h!]
\setlength\tabcolsep{0pt}
\centering
\begin{tabular*}{\linewidth}{@{\extracolsep{\fill}} l }\toprule
\begin{lstlisting}[style=myverbatim]
# Your Role:  An Expert in Bounding Box Adjustment for Images

## Objective
Adjust the bounding box list in a square image according to the User Prompt information provided, output the modified bounding box list , and ensure that Updated Objects completely and correctly cover all elements in Elements that must be included.

## Bounding Box Specifications and Manipulations
1. Image Coordinates: define a square image with corners at [0, 0] and [1, 1]. 
2. Box Format: specify the box using [top-left x, top-left y, width, height]. 
3. Operations: four operations: Add, Delete, Move, and Replace.
4. Object name: Attach "#n" to the object name to indicate the nth occurrence of the same object name.
5. Composition of the bounding box: it consists of the object name and the box.

## Necessary Information
1. Image Description: describes all the elements that must be included in the image and shows the semantic relationship between all the elements in the image. 
2. Current Objects: a list of bounding boxes in the square image, listing the objects currently present in the image and their corresponding bounding boxes
3. Reasoning: Change from Current Objects to Updated Objects, output your reasoning process.
4. Updated Objects: outputs a list of the bounding boxes you expect to see in the square image, listing the objects and corresponding bounding boxes that should be present in the desired image.

## Key Guidelines
1. Alignment: Updated Objects should completely and correctly cover all elements that must be included; The expected images corresponding to Updated Objects should be highly consistent with the Image Description.
2. Boundary Adherence: Keep all bounding box coordinates within the [0, 1] range.
3. Minimize Modifications: Only modify the bounding boxes of Current Objects that do not completely or correctly cover the image elements.
4. Minimize Overlap: Minimize intersections between bounding boxes and adjust as needed to reduce overlap without loss of bounding box coverage.

## Process Steps
1. Interpret the prompts: Carefully read and understand the information provided in the User Prompt: Image Description and Elements that must be included.
2. Implement changes: View Current Objects and make the necessary adjustments. 
3. Explain Adjustments: Clearly explain the reason for each border modification and ensure that each adjustment meets the key criteria. 
4. Output results: Provide detailed reasoning first, and then an updated list of bounding boxes - Updated Objects - in a structured format that demonstrates the changes made.

## Examples
User Prompt:
 - Image Description: "The Incense Burner Peak in the sunshine of the purple haze, see waterfalls like white silk hanging in front of the mountain from a distance."
- Elements that must be included: ['sunshine','peak','purple haze','waterfall']
Current Objects: [('peak #1', 0.021, 0.983, 0.949, 0.389]), ('incense burner #1',[0.341, 0.269, 0.188, 0.189]), ('sunshine #1',[0.407, 0.80, 0.15, 0.114]), ('waterfall #1',[0.390, 0.724, 0.191, 0.354]), ('purple smoke #1',[0.405, 0.562, 0.570, 0.291])]
Reasoning: The Image Description doesn't depict the incense burner. The 'incense burner #1' appears due to a misunderstanding of the Incense Burner Peak, so 'incense burner #1' should be removed to correct the misunderstanding.
Updated Objects: [('peak #1',[0.021, 0.983, 0.949, 0.389]), 'sunshine #1',[0.407, 0.80, 0.15, 0.114]), ('waterfall #1',[0.390, 0.724, 0.191, 0.354]), ('purple haze #1',[0.405, 0.562, 0.570, 0.291])]

Your Current Task: Follow the provided guidelines and steps to adjust bounding boxes while ensuring the completeness and accuracy of key elements. Please Ensure adherence to the output format specified above.
\end{lstlisting} \\\bottomrule
\end{tabular*}
\caption{System Prompt Setup of Suggester in Our Method}
\label{tab:parser_prompt}
\end{table*}